%% file: main.tex
\documentclass{article}
\usepackage{arxiv_preprint,times}

\usepackage{amsmath,amssymb,amsthm,booktabs,graphicx,microtype}
\usepackage{xcolor}
\definecolor{appendixlink}{rgb}{0.16,0.37,0.62}
\usepackage{hyperref,url}
\hypersetup{colorlinks=true,allcolors=black}
\newtheorem{theorem}{Theorem}

\newtheorem{proposition}[theorem]{Proposition}

\newcommand{\norm}[1]{\lVert#1\rVert}
\newcommand{\LSE}{\operatorname{LSE}}
\newcommand{\TV}{\operatorname{TV}}
\title{How Accurate Is Accurate Enough?}
\author{\parbox{\dimexpr\textwidth-2\tabcolsep\relax}{\centering
Ningkang Peng\textsuperscript{1}, Qianfeng Yu\textsuperscript{1}, Jingyang Mao\textsuperscript{1},\\
Xiaoqian Peng\textsuperscript{2}, Yanhui Gu\textsuperscript{1}\\[0.5em]
\normalfont\textsuperscript{1}Nanjing Normal University\\
\textsuperscript{2}Nanjing University of Chinese Medicine\\
\texttt{nkpeng@nnu.edu.cn}, \texttt{gu@njnu.edu.cn}}}
\iclrfinalcopy
\begin{document}
\addtocontents{toc}{\protect\setcounter{tocdepth}{-1}}
\maketitle
\input{sections/abstract}

\suppressfloats[t]
\input{sections/figure1_final}
\input{sections/introduction}
\input{sections/interface}
\input{sections/contracts}
\input{sections/tolerance}
\input{sections/experiments}
\input{sections/discussion}

\input{sections/statements}
\bibliography{references}
\bibliographystyle{state_tolerance_references}
\clearpage
\appendix
\addtocontents{toc}{\protect\setcounter{tocdepth}{2}}
\input{sections/appendix_contents}
\input{sections/appendix_related_work}

\input{sections/appendix_tolerance}
\input{sections/appendix_conditioning}
\input{sections/appendix_primitive}

\input{sections/appendix_foundation}

\input{sections/appendix_learning}
\input{sections/appendix_objectives}

\input{sections/appendix_realization}

\input{sections/appendix_numerics}
\end{document}

%% file: sections/abstract.tex
\begin{abstract}
How accurate must a numerical approximation be within a learning system? Primitive error alone cannot answer this question: errors of the same magnitude can have very different consequences for losses, predictions, and gradients at different learning states. We study this question through the learning objective itself. The objective weights classwise numerical errors nonuniformly according to the current state, so the importance of an error depends not only on its magnitude but also on the class it affects and the weight that class receives. For softmax cross-entropy, we characterize this coupling between class weights and errors and derive the exact extrema of the signed loss change over pairings of fixed non-target probability and score-error multisets, with the target probability and target score error held fixed. Building on this structure, we establish finite-error guarantees that propagate primitive error to losses, probabilities, predictions, and feature gradients, then invert these guarantees to obtain a certified primitive tolerance for the current state under prescribed learning-level error requirements. We give a complete instantiation of the framework in high-dimensional von Mises--Fisher learning. Controlled interventions and a large collection of saved learning states show that identical primitive error can produce substantially different learning consequences, while certified numerical tolerances vary by orders of magnitude across states under the same learning-level requirements. These results show that the adequacy of a numerical approximation must be assessed in relation to the current learning state and the quantity to be preserved; numerical accuracy should itself be treated as part of the learning objective.
\end{abstract}

%% file: sections/figure1_final.tex
\begin{figure}[t]
\centering
\includegraphics[width=\linewidth]{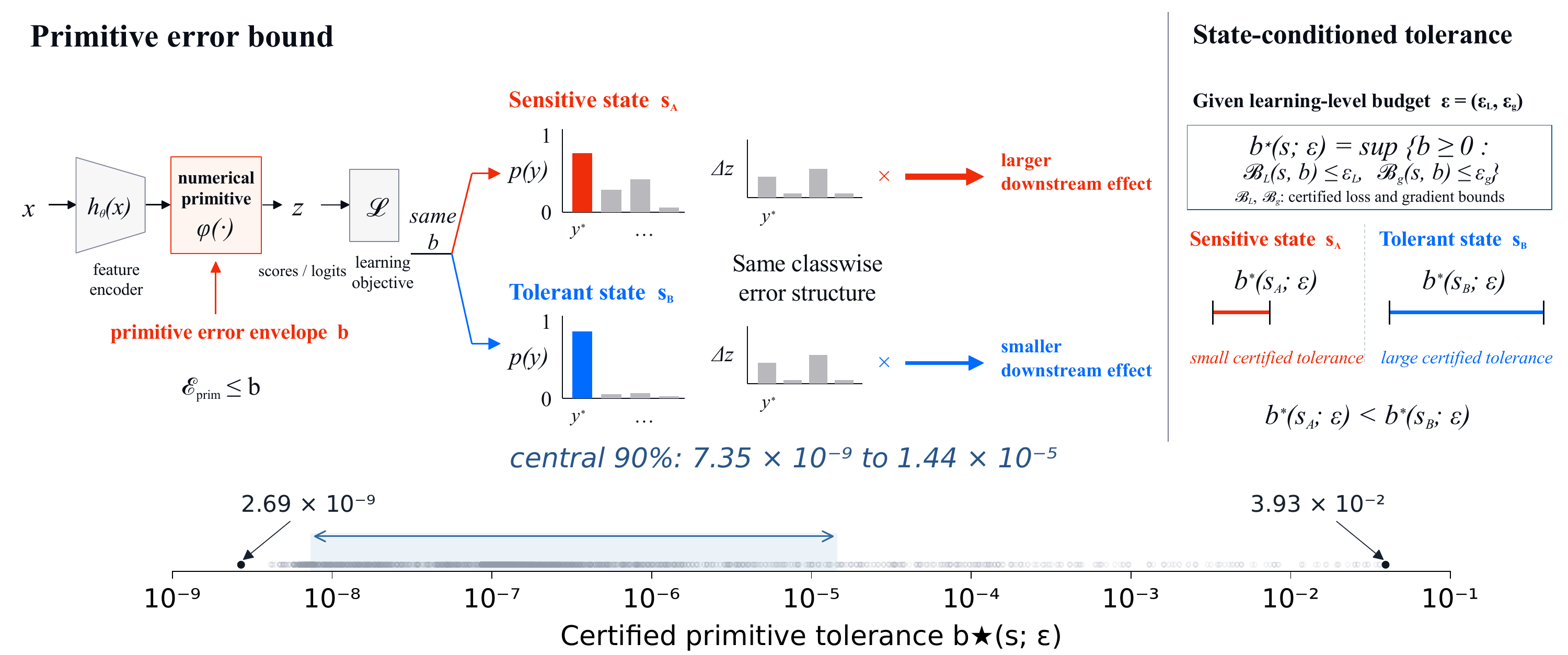}
\caption{From primitive error bounds to state-conditioned numerical tolerance. A primitive error bound propagates through the learning objective. The same classwise error structure can have different downstream effects in different learning states. Inverting certified loss and gradient bounds under budgets $\epsilon_L$ and $\epsilon_g$ yields the certified primitive tolerance $b^\star(s;\boldsymbol\epsilon)$. The upper panels are schematic. The lower panel shows saved-state tolerances under the joint budget $\epsilon_L=\epsilon_g=10^{-5}$.}
\label{fig:concept}
\end{figure}

%% file: sections/introduction.tex
\section{Introduction}
\label{sec:introduction}
Modern machine learning relies on approximate numerical computations, including special functions, normalization constants, integrals, and finite-precision operations \citep{higham2002accuracy,micikevicius2018mixed}. Numerical implementations are commonly assessed by how closely they reproduce a mathematical function. Learning objectives, however, use these values to form scores, probabilities, losses, and gradients. The accuracy required of an intermediate computation therefore depends on how its error affects these downstream quantities \citep{rice1966condition,blanchard2021softmax}.

The same primitive error can have different consequences across learning states. High confidence in the target class can attenuate the effect of a perturbation on the loss or gradient, while competing classes can make the learning signal more sensitive to the same error. Differentiation introduces another source of variation: even when the underlying numerical problem is unchanged, the coordinate scale determines how its error enters the gradient. The error of an approximation and the accuracy required by the current learning state are thus distinct quantities.

We study the second quantity by asking: \textbf{given a learning-level error budget, what primitive error allowance can be certified at the current state?} Starting from requirements on loss, probabilities, prediction stability, or feature gradients, we work backward to the largest allowance accepted by a specified certificate and error family. This state-conditioned tolerance depends on the learning state, the numerical realization, and the downstream requirement.

Our framework propagates primitive errors through the learning objective and inverts the resulting finite-error guarantees. The analysis retains primitive geometry, differentiation scale, and the current probability structure. For softmax cross-entropy, we identify a class-weight--error coupling: even with the target terms and the multisets of non-target probabilities and score errors fixed, changing their classwise pairing can change the loss error. The classical rearrangement inequality gives the exact extrema of the signed loss change. A first-order expansion further describes local certificate sensitivity; tolerances at finite budgets are obtained by nonlinear inversion.

Saved learning states exhibit substantial tolerance variation under common learning-level requirements, including within groups of fixed dimension and temperature. Controlled probability rearrangements isolate the classwise weighting mechanism. Holding reference target probability, entropy, exact top-two margin, geometry, and primitive errors fixed, changing the assignment of non-target probabilities to classes changes actual loss and gradient errors. These aggregate statistics therefore do not fully characterize the interaction between objective weights and numerical error.

Goal-oriented error estimation relates numerical discretization error to a selected quantity of interest and uses this relation to guide adaptive computation \citep{becker2001optimal,giles2002adjoint}. We specialize this perspective to numerical primitives within learning objectives. Current probability weights and class structure yield computable finite-error guarantees that can be inverted into certified primitive tolerances. The cross-entropy coupling result identifies an additional structure of these objectives: the pairing of probability weights and score errors matters even when both multisets are fixed.

High-dimensional von Mises--Fisher learning provides a rigorous instantiation \citep{mardia1999directional,banerjee2005clustering}. Its special-function errors admit explicit bounds, allowing us to follow the full chain from a primitive allowance to a learning-level guarantee and its inverse.

Our contributions are:
\begin{enumerate}
\item We characterize class-weight--error coupling in softmax cross-entropy. With the target terms fixed, changing the pairing of fixed non-target probability and score-error multisets can change the loss error. Proposition~\ref{prop:ce-coupling} gives the exact extrema of the signed loss change.
\item We derive fixed-state loss, probability, prediction, and feature-gradient guarantees and characterize their local sensitivity. Inverting these guarantees yields a certified primitive tolerance under prescribed learning-level budgets.
\item We establish orders-of-magnitude variation in certified tolerances across saved states and isolate classwise weighting through controlled probability rearrangements. At fixed probability summaries, geometry, and primitive errors, class assignments change downstream errors and, under sufficiently strict budgets, numerical adequacy.
\end{enumerate}

%% file: sections/interface.tex
\section{Problem Setup and Instantiation}
\label{sec:interface}
\subsection{Numerical adequacy at a fixed learning state}
A learning state $s$ specifies the inputs of a numerical primitive, the quantities held fixed, and the differentiation coordinate. A realization $\mathcal N$ replaces exact primitive values and derivatives inside an objective. For a specified downstream quantity $G$, let $E_{\mathcal N}^G(s)$ denote the induced absolute error, probability distance, or vector-norm error. Given a budget $\epsilon_G$, numerical adequacy means $E_{\mathcal N}^G(s)\le\epsilon_G$.

A certificate $B_{\mathcal N}^G(s)$ supplies a computable sufficient condition,
\begin{equation}
E_{\mathcal N}^G(s)\le B_{\mathcal N}^G(s)\le\epsilon_G.
\label{eq:adequacy-interface}
\end{equation}
The certificate is computed from approximate outputs and valid primitive error allowances. Acceptance guarantees adequacy; otherwise, the bound leaves adequacy unresolved. Inverting the certificate therefore yields the \emph{largest allowance certified by the chosen bound}, which may be smaller than the allowance that the actual learning error would permit.

\subsection{A paired primitive for vMF learning}
For feature dimension $p\ge5$, let $\nu=p/2-1$. Define
\begin{equation}
\Phi_\nu(x)=\log I_\nu(x)-\nu\log x,\qquad
R_\nu(x)=I_{\nu+1}(x)/I_\nu(x),\qquad \Phi_\nu'=R_\nu.
\label{eq:primitive-instance}
\end{equation}
The ratio is central to directional inference and vMF parameter estimation \citep{sra2012vmf}. The functions use their continuous extensions at $x=0$. We consider a differentiable paired realization $(F,A)$ with $F'=A$ and $A(0)=0$. Pairing makes the approximate objective and its feature derivative refer to the same mathematical surrogate.

Let $\mu_j\in\mathbb R^p$ be unit class directions, $k_j\ge0$ concentrations, $c_j$ finite shared class biases, $\tau>0$ a temperature, and $f\in\mathbb R^p$ a feature of arbitrary norm. With finitely many classes $C\ge2$, define
\begin{align}
v_j&=k_j\mu_j+f/\tau, & r_j&=\norm{v_j},\\
q_j&=\Phi_\nu(r_j)-\Phi_\nu(k_j)+c_j,&
\widehat q_j&=F(r_j)-F(k_j)+c_j.
\label{eq:learning-instance}
\end{align}
Write $P=\operatorname{softmax}(q)$, $\pi=\widehat P=\operatorname{softmax}(\widehat q)$, and $L_y=\LSE(q)-q_y$, with $\widehat L_y$ defined analogously. The log-partition and log-sum-exp structure connects probabilities to derivatives \citep{wainwright2008graphical,boyd2004convex}. Class statistics, biases, and temperature remain fixed in the Euclidean feature derivative $\nabla_f$.

These normalizer differences occur in probabilistic contrastive objectives \citep{du2024proco,he2025patt}. Supervised probabilistic vMF embeddings provide another instance, with an additional ratio-dependent attraction term \citep{Scott_2021_ICCV}; its propagation is given in Appendix~\ref{app:objectives}.

Our experiments instantiate $(F,A)$ by the stable second-order pair $(F_{2,\nu},A_{2,\nu})$. It satisfies $F_{2,\nu}'=A_{2,\nu}$ and admits both a uniform ratio allowance $(4\nu^3)^{-1}$ and a tighter state-dependent allowance. The construction, stable endpoint formula, and full special-function proofs are in Appendices~\ref{sec:primitive}, \ref{app:proof}, and~\ref{app:realization}. The analysis below applies to any paired realization with valid error allowances.

%% file: sections/contracts.tex
\section{From Primitive Error to Learning Consequences}
\label{sec:contracts}
\subsection{Score allowances and a geometry-only contract}
\label{sec:contrastive-contracts}
Suppose a finite integrable allowance satisfies $|R_\nu(x)-A(x)|\le e(x)$ on every integration path between $k_j$ and $r_j$. Pairing implies
\begin{equation}
|q_j-\widehat q_j|\le d_j:=\int_{\min(k_j,r_j)}^{\max(k_j,r_j)}e(x)\,dx,
\qquad |R_\nu(r_j)-A(r_j)|\le a_j.
\label{eq:score-allowance-main}
\end{equation}
Here $a_j$ is an unscaled ratio allowance. Let $d=\max_jd_j$.
\begin{proposition}[Fixed-state contrastive contract]
\label{prop:contrastive-contract}
For the state and paired realization in Section~\ref{sec:interface},
\begin{align}
|\widehat L_y-L_y|&\le d_y+d, & \TV(P,\pi)&\le\tanh(d/2),\\
\norm{\nabla_f\widehat L_y-\nabla_fL_y}
&\le \frac{2\tanh(d/2)+\max_j a_j+a_y}{\tau}.
\label{eq:base-contract-main}
\end{align}
If the approximate top-two score margin $\widehat m$ exceeds $2d$, the exact and approximate predictions agree. The same sufficient condition applied to the exact margin is valid for reference-side evaluation.
\end{proposition}
The bounds have no explicit class-count multiplier. They retain endpoint geometry and the derivative scale $1/\tau$, but largely discard the current probability structure. Proofs appear in Appendix~\ref{app:learning}.

\subsection{Retaining the current probability structure}
\label{sec:objective-conditioned-contract}
Define
\begin{equation}
S_j^\pm=\sum_{i\ne j}\pi_i\{\exp[\pm(d_i+d_j)]-1\},\qquad
\ell_j=\frac{\pi_j}{1+S_j^+},\quad
u_j=\min\!\left\{1,\frac{\pi_j}{1+S_j^-}\right\}.
\label{eq:probability-interval-main}
\end{equation}
The denominators remain positive because the class-$j$ self-term has no error. These intervals are computed from the approximate probabilities and classwise score allowances; the loss bound below also uses the target identity.
\begin{proposition}[Objective-conditioned contract]
\label{prop:objective-conditioned-contract}
The following quantities bound loss error, probability TV, and feature-gradient error, respectively:
\begin{align}
B_L&=\max\{\log(1+S_y^+),-\log(1+S_y^-)\},\\
B_P&=\min\!\left\{\tanh(d/2),\frac12\sum_j\max(\pi_j-\ell_j,u_j-\pi_j)\right\},\\
B_g&=\frac{2B_P+\sum_{j\ne y}\pi_j a_j+(1-\pi_y)a_y}{\tau}.
\label{eq:objective-contract-main}
\end{align}
Each is no larger than the corresponding bound in Proposition~\ref{prop:contrastive-contract}.
\end{proposition}
\paragraph{Class-weight--error coupling in cross-entropy.}
For finite reference logits $q$, let $\widehat q=q+e$ and $P=\operatorname{softmax}(q)$. The signed loss change satisfies the exact identity
\begin{equation}
\Delta L:=L_y(\widehat q)-L_y(q)
=\log\sum_jP_j\exp(e_j)-e_y.
\label{eq:ce-error-identity}
\end{equation}
\begin{proposition}[Class-weight--error coupling]
\label{prop:ce-coupling}
Fix $t=P_y\in(0,1)$ and $e_y\in\mathbb R$. Let $p_1\le\cdots\le p_m$ be positive non-target probabilities summing to $1-t$, and $a_1\le\cdots\le a_m$ finite non-target score errors, with $m=C-1$. Over all pairings of these two multisets, the signed loss change has attained extrema
\begin{align}
D_-&=\log\!\left[t+\sum_{i=1}^mp_i\exp(a_{m+1-i}-e_y)\right],\\
D_+&=\log\!\left[t+\sum_{i=1}^mp_i\exp(a_i-e_y)\right].
\end{align}
Consequently $\max_\sigma|\Delta L_\sigma|=\max\{|D_-|,|D_+|\}$.
\end{proposition}
These extrema follow from the classical rearrangement inequality: matching larger probabilities with larger signed errors maximizes the signed loss change, while reverse matching minimizes it. The smallest absolute error can occur at an intermediate pairing. The result applies to any softmax cross-entropy objective; Appendix~\ref{app:ce-coupling-proof} gives the proof. Gradient-error norms also depend on class directions.

To see the gradient mechanism, put $h_j=R_\nu(r_j)v_j/r_j$ and $\widehat h_j=A(r_j)v_j/r_j$, using zero at $r_j=0$. Then
\begin{equation}
\tau(\nabla_f\widehat L_y-\nabla_fL_y)
=\sum_j(\pi_j-P_j)h_j+
\sum_j(\pi_j-\mathbf1_{j=y})(\widehat h_j-h_j).
\label{eq:gradient-mechanism-main}
\end{equation}
The first term captures probability redistribution; the second weights each primitive derivative error by the current objective, with weights $\pi_j$ for non-target classes and $1-\pi_y$ for the target. Each error is weighted according to its role in the current objective. Changing the correspondence between probability weights and classwise errors can therefore change loss and gradient errors while preserving target probability and both multisets. For gradients, class directions also determine how these weighted errors combine. Appendix~\ref{app:probability-conditioned} gives the proof and a linear-in-$C$ implementation.

\paragraph{Primitive accuracy and learning adequacy.}
Common-shift cancellation illustrates why ranking primitive errors can give a different ordering from ranking learning errors. Consider two paired perturbations of the exact potential in a two-class example:
\begin{center}
\fbox{\begin{minipage}{0.94\linewidth}\small
A: ratio error $\varepsilon$; score shifts $(\varepsilon/2,5\varepsilon/6)$; CE changes.\\
B: ratio error $2\varepsilon$; score shifts $(2\varepsilon,2\varepsilon)$; CE is unchanged.
\end{minipage}}
\end{center}
The larger error cancels as a common score shift, whereas the smaller, nonuniform error changes the loss. Appendix~\ref{app:ranking} gives the paired construction and a budget separating their adequacy.

%% file: sections/tolerance.tex
\section{From Learning Budgets to State-Conditioned Tolerance}
\label{sec:tolerance}
\subsection{Inverting a fixed-state contract}
Let $g_j=|r_j-k_j|$ and consider a scalar primitive allowance $b\ge0$ that is valid along every endpoint integration path. At a fixed state, set
\begin{equation}
d_j(b)=b g_j,\qquad a_j(b)=b.
\label{eq:uniform-allowance-family}
\end{equation}
Fix the numerical realization $\mathcal N$, the propagation certificate $\mathcal B$, and the allowance family $\mathcal E_s(b)$ in Eq.~\eqref{eq:uniform-allowance-family}. Keep the approximate scores and probabilities produced by $\mathcal N$ fixed while varying this allowance. Substitution into Proposition~\ref{prop:objective-conditioned-contract} gives $B_L(s,b)$, $B_P(s,b)$, and $B_g(s,b)$.

For any selected set of downstream quantities $\mathcal G\subseteq\{L,P,g\}$ and positive budgets $\boldsymbol\epsilon$, define
\begin{equation}
b_{\star,\mathcal N,\mathcal B,\mathcal E}(s;\boldsymbol\epsilon)
=\sup\{b\ge0: B_G(s,b)\le\epsilon_G\ \text{for all }G\in\mathcal G\}.
\label{eq:state-tolerance}
\end{equation}
With $\mathcal N$, $\mathcal B$, and $\mathcal E$ fixed, we abbreviate this threshold as $b_\star(s;\boldsymbol\epsilon)$. It is the largest uniform primitive allowance accepted by this certificate for the current numerical outputs, providing post-evaluation certification of the chosen realization. A valid primitive envelope $b\le b_\star$ certifies the requested learning quantities. The primitive allowance must hold over the full endpoint integration paths.

\begin{proposition}[Monotone inversion]
\label{prop:inversion}
At a finite fixed state, the functions $B_L(s,b)$, $B_P(s,b)$, and $B_g(s,b)$ are continuous and nondecreasing on $b\ge0$, and vanish at zero. Their simultaneous accepted set is an interval starting at zero. Any finite threshold can therefore be found by bracketed scalar inversion without access to the exact learning outputs.
\end{proposition}
For prediction stability, add $2b\max_jg_j<\widehat m$. The strict inequality specifies the open acceptance boundary of the margin test. Probabilities and losses can be computed in the log domain during inversion. Appendix~\ref{app:tolerance-proof} proves the monotonicity and gives the numerical protocol.

\subsection{First-order state sensitivity}
The first-order expansion describes local behavior as $b\to0$ at a fixed state. When probabilities are highly saturated and local sensitivity is close to zero, higher-order terms can substantially affect the inverse at finite budgets. We use the full nonlinear bound for all reported tolerances. Define
\begin{align}
c_L(s)&=\sum_{j\ne y}\pi_j(g_j+g_y),\label{eq:sensitivity-loss}\\
c_P(s)&=\min\!\left\{\frac{\max_jg_j}{2},\sum_j\pi_j(1-\pi_j)g_j\right\},\\
c_g(s)&=\frac{2c_P(s)+2(1-\pi_y)}{\tau}.
\label{eq:first-order-coefficients}
\end{align}
\begin{proposition}[First-order sensitivity]
\label{prop:sensitivity}
At a fixed finite state, as $b\downarrow0$,
\begin{equation}
B_L(s,b)=c_L(s)b+O(b^2),\quad
B_P(s,b)=c_P(s)b+O(b^2),\quad
B_g(s,b)=c_g(s)b+O(b^2).
\end{equation}
Thus, in the small-budget regime, the joint loss--gradient threshold is approximated by
\begin{equation}
b_\star(s;\epsilon_L,\epsilon_g)
\approx \min\!\left\{\frac{\epsilon_L}{c_L(s)},\frac{\epsilon_g}{c_g(s)}\right\},
\label{eq:first-order-tolerance}
\end{equation}
with an inactive zero-sensitivity constraint omitted.
\end{proposition}
Target probability enters the gradient sensitivity through $1-\pi_y$, while probability TV retains the full distribution and endpoint geometry. Temperature appears explicitly through the differentiation scale. Feature norm constrains the endpoint geometry through $g_j\le\norm f/\tau$. The top-two margin controls prediction stability, whereas target-loss sensitivity also depends on the identity of the dominant class.

These coefficients separate two ways in which a class can matter. In $c_L$, an endpoint gap is weighted by the probability assigned to a competing class. A large gap for a nearly inactive competitor contributes little, while the same gap matters more when that competitor carries appreciable probability. The target gap is multiplied by the total non-target mass. Thus high target confidence can relax the required accuracy without changing the primitive geometry.

The gradient coefficient also contains the probability-redistribution term $c_P$. Two states with the same target probability can distribute the remaining mass differently across endpoint gaps, and therefore have different gradient sensitivities. The explicit factor $1/\tau$ converts this effect into the chosen feature coordinate. Changing temperature with $f$ fixed also changes geometry and probabilities; the clean coordinate control instead holds $f/\tau$ fixed. Taken together, these terms explain why target confidence, temperature, or feature norm alone cannot determine numerical tolerance. The inverse contract combines their effects under the requested learning budget.

%% file: sections/experiments.tex
\section{Experiments}
\label{sec:experiments}
We sample 5,120 query-level learning states from feature records at existing training checkpoints and saved training batches, retaining the complete class statistics and numerical geometry associated with each query. High-precision reference calculations and the full reproducibility protocol are described in Appendices~\ref{app:tolerance-protocol} and~\ref{app:numerics-reference}.

\subsection{Different learning consequences of the same primitive error}
\label{sec:exp-conditioning}
\begin{figure}[t]
\centering
\includegraphics[width=\linewidth]{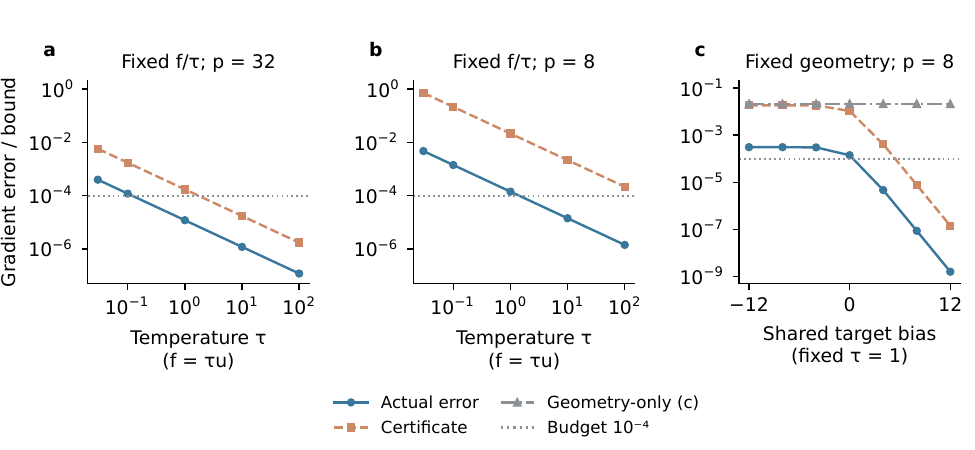}
\caption{Learning consequences change while primitive error stays fixed. \textbf{a,b,} At fixed $f/\tau$, changing temperature rescales feature-gradient error without changing the scalar numerical problem. \textbf{c,} A shared target-class bias changes objective weights at fixed geometry and temperature. Blue shows actual error; orange shows the geometry-only certificate in a,b and the objective-conditioned certificate in c. The gray curve in c is the unchanged geometry-only bound; dotted lines mark a common gradient budget.}
\label{fig:conditioning}
\end{figure}

We first separate two mechanisms by controlled intervention. For coordinate conditioning, keep $u=f/\tau$, class statistics, and biases fixed while setting $f_\tau=\tau u$. All scalar vMF inputs and numerical errors remain identical. The feature-gradient discrepancy nevertheless scales as $1/\tau$, crossing a fixed absolute gradient budget (Fig.~\ref{fig:conditioning}). This change follows directly from chain-rule scaling in the chosen feature coordinate: absolute gradient error varies with coordinate scale while the underlying numerical problem remains fixed.

For objective conditioning, add the same target-class bias to exact and approximate scores, keeping primitive geometry, temperature, the score-error vector, and primitive allowances fixed. Changing target probability changes how the loss weights the same numerical perturbation. The objective-conditioned bound follows this change, whereas the geometry-only bound remains fixed. The two controls isolate distinct mechanisms: coordinate scaling changes the units of an absolute gradient budget, while objective conditioning changes the weights applied to fixed classwise errors. Figure~\ref{fig:conditioning} compares their effects; Appendix~\ref{app:conditioning-detail} defines the interventions.

\subsection{Tolerance distributions in saved states}
\label{sec:exp-tolerance}
\begin{figure}[t]
\centering
\includegraphics[width=\linewidth]{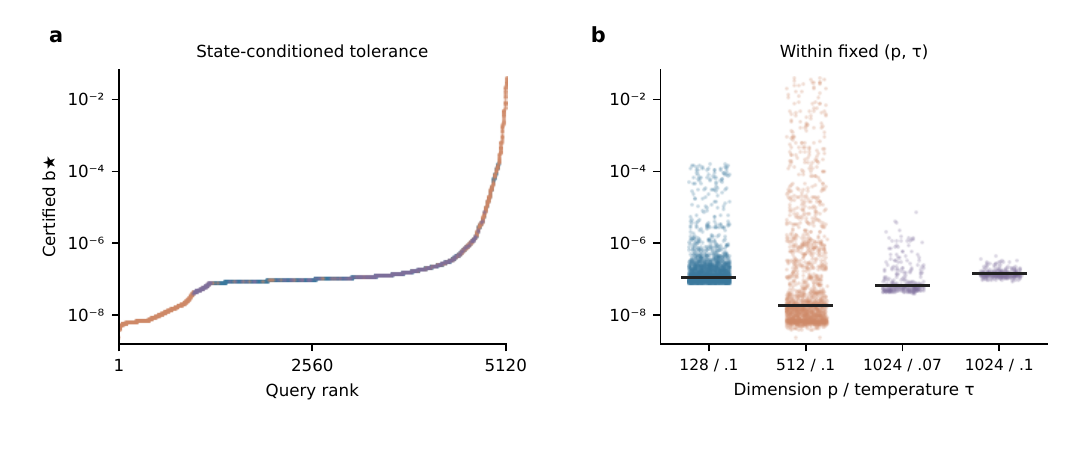}
\caption{Query-level tolerance varies under a common learning-error budget. \textbf{a,} Ordered joint tolerances at budget $10^{-5}$. \textbf{b,} The same queries grouped by dimension and temperature; black bars denote medians. Colors indicate dimension: blue 128, orange 512, and purple 1024.}
\label{fig:tolerance-distribution}
\end{figure}

We compare numerical requirements across saved queries under common absolute loss and feature-gradient error budgets. At the joint budget $10^{-5}$, the central 90\% of certified primitive tolerances spans 3.29 orders of magnitude and the full range spans 7.16. At the looser budget $10^{-4}$, the corresponding spans are 3.29 and 6.24 orders of magnitude. Substantial variation remains within the fixed $p=512$, $\tau=0.1$ group. These descriptive query-level distributions show heterogeneity at both budgets in the saved collection; Figure~\ref{fig:tolerance-distribution} displays the distributions and Table~\ref{tab:tolerance} reports their quantiles.

Under a prespecified paired perturbation family, detected actual gradient-response thresholds at budget $10^{-5}$ also span multiple orders of magnitude: 6.35 overall and 5.70 within the fixed $p=512$, $\tau=0.1$ group. The sufficient certificate is typically more conservative along this fixed perturbation family; Appendix~\ref{app:actual-response} describes the response check.
\begin{table}[t]
\centering
\caption{Distribution of certified primitive tolerance under loss and joint budgets. Columns report the 5th percentile, median, 95th percentile, and full logarithmic span $\log_{10}(b_{\max}/b_{\min})$.}
\label{tab:tolerance}
\begin{tabular}{llrrrr}
\toprule
Budget & Quantity & 5\% & Median & 95\% & Full span\\
\midrule
$10^{-4}$ & Loss & $4.86\!\times\!10^{-7}$ & $9.58\!\times\!10^{-6}$ & $2.40\!\times\!10^{-3}$ & 5.74\\
$10^{-4}$ & Joint & $7.35\!\times\!10^{-8}$ & $1.06\!\times\!10^{-6}$ & $1.44\!\times\!10^{-4}$ & 6.24\\
$10^{-5}$ & Loss & $4.86\!\times\!10^{-8}$ & $9.58\!\times\!10^{-7}$ & $2.86\!\times\!10^{-4}$ & 6.65\\
$10^{-5}$ & Joint & $7.35\!\times\!10^{-9}$ & $1.06\!\times\!10^{-7}$ & $1.44\!\times\!10^{-5}$ & 7.16\\
\bottomrule
\end{tabular}
\end{table}
\subsection{How do error structure and objective weights jointly determine tolerance?}
\label{sec:exp-factors}
\begin{figure}[t]
\centering
\includegraphics[width=\linewidth]{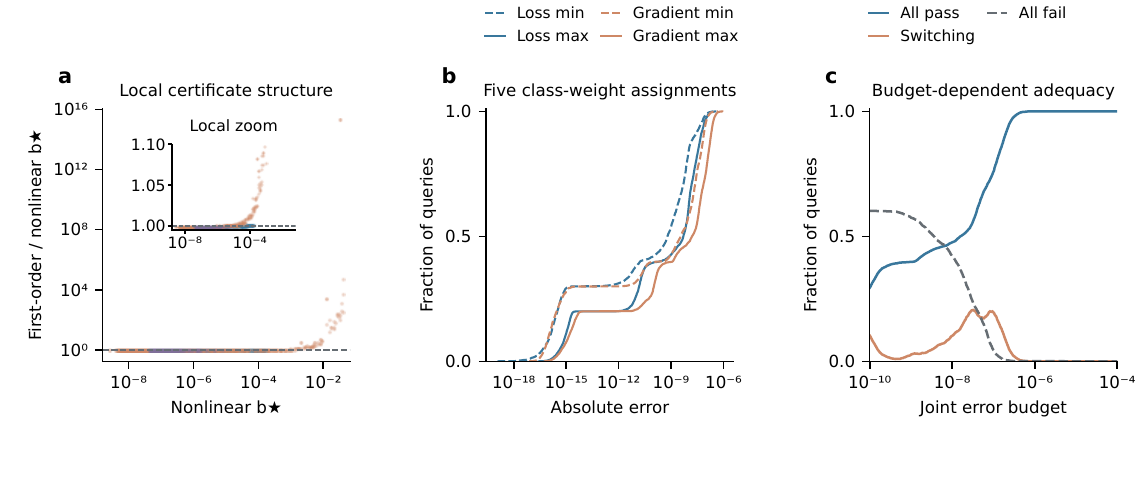}
\caption{Local sensitivity and classwise correspondence shape numerical requirements. \textbf{a,} Ratio of first-order to nonlinear tolerance at budget $10^{-5}$; the inset magnifies the region near one. Colors denote dimension as in Fig.~\ref{fig:tolerance-distribution}. \textbf{b,} Cumulative distributions of minimum and maximum absolute errors over five fixed assignments of non-target probabilities. Target probability, entropy, exact top-two margin, geometry, and primitive errors are held fixed. \textbf{c,} Fractions of queries for which all assignments satisfy the joint budget, assignments yield different decisions, or all assignments fail.}
\label{fig:first-order-prediction}
\end{figure}

The first-order expansion in Eqs.~\eqref{eq:sensitivity-loss}--\eqref{eq:first-order-tolerance} characterizes local structure as $b\to0$. At the joint budget $10^{-5}$, the first-order prediction differs from the nonlinear tolerance by at most $1\%$ for 4,989 of the 5,120 queries ($97.44\%$), where relative deviation is measured against the nonlinear tolerance. In highly saturated probability states, however, local sensitivity approaches zero and higher-order terms can become important before the linear inverse reaches the prescribed budget. Figure~\ref{fig:first-order-prediction}a compares the first-order and nonlinear inverses, with an inset around agreement. All reported tolerances are obtained from the full nonlinear inversion.

The first-order expansion shows that sensitivity depends on the joint action of probability weights, endpoint geometry, and differentiation scale. To directly test the role of classwise probability--error correspondence, we construct controlled probability distributions on the saved query geometries, set the target probability to $0.5$, and apply five prespecified rearrangements of the non-target probabilities. Across rearrangements, the probability multiset, entropy, exact top-two margin, and underlying numerical errors remain fixed; only the assignment of probability weights to classes changes. Shared additive biases in the exact and approximate logits realize these assignments without changing the classwise numerical errors.

Actual loss and gradient errors vary with the class assignment at fixed aggregate statistics (Fig.~\ref{fig:first-order-prediction}b), as predicted by Proposition~\ref{prop:ce-coupling} and the gradient decomposition in Eq.~\eqref{eq:gradient-mechanism-main}. Figure~\ref{fig:first-order-prediction}c places these changes on the budget scale: all five assignments satisfy joint budgets of $10^{-4}$ and $10^{-5}$, whereas 1,001 queries change adequacy decisions across assignments at $10^{-7}$. Classwise correspondence thus affects actual learning errors, with its impact on adequacy depending on the required error scale. Appendix~\ref{app:fixed-target-permutation} gives the rearrangement definitions and complete budget results.

%% file: sections/discussion.tex
\section{Discussion and Scope}
\label{sec:discussion}
The framework connects primitive error to the accuracy of specified learning quantities. Forward propagation gives finite-error guarantees, and inversion turns a prescribed budget into a certified allowance at the current state. The vMF instance makes this chain explicit through a paired special-function realization. Appendix~\ref{app:related-work} relates the analysis to numerical precision, inexact optimization, and probabilistic representation learning.

\paragraph{What the inverse guarantee means.}
The inverted threshold provides a state-dependent sufficient guarantee for the current numerical outputs, chosen certificate, and primitive-error family. Measured response thresholds also vary substantially across states and are typically larger than the certified allowances along the fixed perturbation family. Classwise correlations and cancellation can keep actual errors within budget even when the certificate is inconclusive. Each evaluator is assessed through its own primitive bounds and approximate probabilities.

\paragraph{Specifying learning-error budgets.}
We treat learning-level error budgets as external specifications determined by the application. A loss-error budget may reflect a numerical stability requirement, while a gradient-error budget may be set relative to the current gradient norm, the optimization step size, or an allowed update perturbation. Our analysis addresses propagation and inversion for a given budget; deriving such budgets automatically from long-term optimization or task-level requirements remains a direction for future work.

\paragraph{From feature gradients to parameter gradients.}
For the same differentiable encoder $f=h_\theta(x)$ at a fixed input and parameter state, let $J_\theta=\partial h_\theta(x)/\partial\theta$. When the numerical discrepancy enters through the loss as a function of $f$, the shared encoder Jacobian propagates the feature-gradient error to parameter space:
\[
\nabla_\theta L=J_\theta^\top\nabla_f L,
\qquad
\|\Delta\nabla_\theta L\|_2
\le \|J_\theta\|_{2\to2}\,\|\Delta\nabla_f L\|_2.
\]
Thus the feature-gradient certificate controls the error passed to the encoder, with parameter-gradient sensitivity determined by its Jacobian at the current state.

\paragraph{Mathematical and execution scope.}
The theory characterizes numerical quantities at fixed learning states in specified differentiation coordinates. Propagating these guarantees to long-term optimization trajectories and task-level performance is a direction for further study. The certificates control the mathematical error of the analytic surrogate. Full floating-point execution guarantees would additionally require explicit allowances for rounding, endpoint construction, transcendental evaluation, and numerical reductions. High-precision comparisons assess the implementation over the tested range.

State-conditioned certificates also support coarse-to-fine analysis: an inexpensive bound is evaluated first, followed by a tighter bound when needed to resolve adequacy. This procedure gives the same decisions as always evaluating the tighter bound (Appendix~\ref{sec:refinement}).

\paragraph{Beyond the vMF instance.}
The paired vMF primitive supplies one explicit setting in which both primitive allowances and their learning-level propagation are available. Applying the interface to a different numerical primitive requires its own valid allowances and propagation analysis. More broadly, the framework connects requirements on underlying numerical errors to the quantities that a learning objective needs to protect.

%% file: sections/statements.tex
\section*{Reproducibility Statement}
The appendix gives the primitive and learning-contract proofs, tolerance inversion, controlled interventions, and shared-geometry reference protocol. The source materials provide the saved-query data, numerical settings, and verification records. The analysis covers 5,120 queries, with queries from shared training runs treated as dependent observations. Mathematical guarantees and floating-point evaluation are distinguished throughout.

\section*{AI Use Statement}
AI tools were used to assist with manuscript language polishing, code drafting, mathematical derivations, and figure design. The authors take full responsibility for all content.

%% file: sections/appendix_contents.tex
\begin{center}
{\Large\bfseries How Accurate Is Accurate Enough?\par}
\vspace{0.6em}
{\Large Supplementary Material\par}
\end{center}
\vspace{0.5em}
\begingroup
\hypersetup{linkcolor=appendixlink,linktoc=section}
\makeatletter
\renewcommand{\l@section}[2]{\addpenalty{\@secpenalty}\addvspace{0.45em}{\bfseries\@dottedtocline{1}{0em}{2em}{#1}{#2}}}
\renewcommand{\l@subsection}{\@dottedtocline{2}{2em}{2.6em}}
\makeatother
\tableofcontents
\endgroup
\clearpage

%% file: sections/appendix_related_work.tex
\section{Related Work}
\label{app:related-work}
\paragraph{Numerical precision in learning.}
Mixed-precision training uses lower-precision arithmetic while preserving training performance through techniques such as loss scaling and higher-precision weight updates \citep{micikevicius2018mixed}. Special-function algorithms provide tools for accurate evaluation across parameter regimes \citep{gil2007numerical}, while rounding-error analysis of softmax and log-sum-exp explains how equivalent formulas differ in floating-point stability \citep{blanchard2021softmax}. These perspectives motivate careful numerical realization inside learning systems. We study the complementary question of how much primitive error a particular learning state can tolerate under a specified downstream requirement, with tolerance depending on the objective weights, geometry, and differentiation scale.

\paragraph{Inexact optimization and goal-oriented error analysis.}
Gradient methods with errors and inexact-oracle methods relate controlled function or gradient inaccuracies to optimization behavior \citep{bertsekas2000errors,devolder2014inexact}. Goal-oriented error estimation uses sensitivity and duality to relate local errors to a selected output and guide adaptive computation \citep{becker2001optimal,giles2002adjoint}. Our contribution lies within this broader objective-dependent view: for a numerical primitive embedded in a probabilistic learning objective, we construct computable finite-error guarantees for losses, probabilities, predictions, and feature gradients, and invert them to obtain a sufficient primitive tolerance at the current state. This fixed-state analysis provides primitive-derived error bounds of the kind that inexact optimization takes as inputs; propagating them along an optimization trajectory requires further assumptions.

\paragraph{von Mises--Fisher primitives in probabilistic representation learning.}
ProCo models class-conditional features with von Mises--Fisher distributions for probabilistic contrastive learning \citep{du2024proco}; probabilistic vMF embeddings also support supervised classification and retrieval \citep{Scott_2021_ICCV}. Their normalizing constants involve modified Bessel functions, and differentiation introduces Bessel ratios, connecting probabilistic scores and gradients to special-function evaluation. High-dimensional vMF learning provides an explicit instance in which primitive approximation errors can be rigorously bounded and propagated to the learning quantities of interest.

%% file: sections/appendix_tolerance.tex
\section{Tolerance Inversion, Sensitivity, and Refinement}
\label{app:tolerance-proof}
\subsection{Monotonicity of the inverse contract}
Fix finite approximate logits and hence $0<\pi_j<1$, endpoint gaps $g_j\ge0$, a target, and $\tau>0$. With $d_j=bg_j$, each $S_j^+$ is continuous and nondecreasing in $b$, while $S_j^-$ is continuous and nonincreasing. Consequently $\log(1+S_y^+)$ and $-\log(1+S_y^-)$ are nondecreasing. The lower probability endpoint decreases and the upper endpoint increases; clipping the latter at one preserves validity and monotonicity. Both candidate TV bounds are nondecreasing, so their minimum is nondecreasing as well. Finally, the ratio contribution to $B_g$ is $2b(1-\pi_y)/\tau$. These facts prove Proposition~\ref{prop:inversion}. All bounds vanish at zero. Their sublevel sets are closed intervals starting at zero, possibly unbounded if the corresponding constraint is inactive. Intersecting the selected sublevel sets gives Eq.~\eqref{eq:state-tolerance}. The strict prediction-margin condition has an open upper boundary.

\subsection{Derivation of the first-order coefficients}
Let $t_j=\sum_{i\ne j}\pi_i(g_i+g_j)$. Taylor expansion gives $S_j^\pm=\pm b t_j+O(b^2)$, and hence
\[
\log(1+S_y^+)=b t_y+O(b^2),\qquad
-\log(1+S_y^-)=b t_y+O(b^2).
\]
This yields $c_L=t_y$. For sufficiently small $b$ at a fixed finite state, the probability clipping is inactive, and both coordinate deviations have expansion $b\pi_jt_j+O(b^2)$. Their half-sum has coefficient
\[
\frac12\sum_j\pi_jt_j=\sum_j\pi_j(1-\pi_j)g_j.
\]
The other TV bound, $\tanh(b\max_jg_j/2)$, has coefficient $\max_jg_j/2$. Taking the minimum gives $c_P$, including the case where the coefficients coincide. The unscaled ratio contribution is exactly $2b(1-\pi_y)$, yielding $c_g$. These expansions prove Proposition~\ref{prop:sensitivity}. A positive first-order coefficient gives an inverse threshold $\epsilon/c+O(\epsilon^2)$ as its budget decreases; simultaneous small budgets select the active constraint. When a coefficient vanishes, the full nonlinear bound determines whether higher-order terms constrain the tolerance or the constraint remains inactive.

\subsection{Decision equivalence under certificate refinement}
\label{sec:refinement}
For the same numerical realization, let $\mathbf B^{\rm cheap}$ be an inexpensive certificate and $\mathbf B^{\rm tight}$ a tighter certificate, satisfying
\[
\mathbf E\le\mathbf B^{\rm tight}\le\mathbf B^{\rm cheap}.
\]
The cascade first tests $\mathbf B^{\rm cheap}\le\boldsymbol\epsilon$ and computes $\mathbf B^{\rm tight}$ only when the first test is inconclusive. Both stages certify the same evaluator outputs through the same downstream propagation, using successively tighter primitive envelopes.

The ordering above ensures that acceptance by the inexpensive bound implies acceptance by the tighter bound. Otherwise, the cascade evaluates the tighter bound directly. The cascade and the policy that always evaluates the tighter bound therefore return the same decision at every state. 

\subsection{Saved states and inversion protocol}
\label{app:tolerance-protocol}
The saved-query analysis uses the same complete class banks and numerical geometry as the reference comparisons in Appendix~\ref{app:numerics}. At fixed approximate probabilities, we set $d_j=b|r_j-k_j|$ and $a_j=b$ and invert the loss and gradient certificates by bracketed bisection. Stable log-domain reductions evaluate the probability intervals over wide allowance ranges. The first-order prediction uses the same states and budgets without fitted coefficients.

The nonlinear inverse remains well defined when saturated probabilities make the first-order coefficients small. In this regime, higher-order terms can restrict the accepted allowance far earlier than a linear extrapolation predicts. Figure~\ref{fig:first-order-prediction}a shows both the agreement of the inverses for most queries and their large departures in highly saturated states.

\subsection{A paired counterexample to primitive-only ranking}
\label{app:ranking}
Take $p=8$, $\tau=1$, $f=e_1$, two class directions equal to $e_1$, and concentrations $(1,2)$, so the query radii are $(2,3)$. Let the target be the first class and $\varepsilon=10^{-4}$. Define paired potentials using the exact potential $\Phi$:
\[
F^{\mathrm A}(x)=\Phi(x)+\varepsilon H(x),\qquad
F^{\mathrm B}(x)=\Phi(x)+2\varepsilon G(x),
\]
where
\[
H(x)=\begin{cases}x^2/6,&0\le x\le3,\\x-3/2,&x\ge3,\end{cases}
\qquad
G(x)=\begin{cases}x^2/2,&0\le x\le1,\\x-1/2,&x\ge1.\end{cases}
\]
Both potentials are continuously differentiable and use their derivatives as ratio realizations. Their derivative errors vanish at zero and attain global maxima $\varepsilon$ and $2\varepsilon$. At the query, score errors are $(\varepsilon/2,5\varepsilon/6)$ for A and $(2\varepsilon,2\varepsilon)$ for B. B leaves cross-entropy exactly unchanged by common-shift invariance; A has loss error approximately $1.7568\times10^{-5}$. Thus B meets a $10^{-5}$ loss budget while A does not, despite A having a smaller maximum primitive error.

Common-shift cancellation thus allows a larger primitive error to produce a smaller loss error.

%% file: sections/appendix_conditioning.tex
\section{Conditioning Controls and Class-Weight Rearrangement}
\label{app:conditioning-detail}
The coordinate controls in Figure~\ref{fig:conditioning}a,b fix $u=f/\tau$, class directions, concentrations, and biases, then set $f_\tau=\tau u$. Every scalar vMF input and error is unchanged within each dimension. The feature-gradient error and its certificate scale as $1/\tau$, so the control isolates the effect of the differentiation coordinate on an absolute budget.

The objective control in Figure~\ref{fig:conditioning}c adds the same target-class bias to exact and approximate logits. Geometry, temperature, score errors, ratio errors, and primitive allowances remain fixed. Changing only the softmax weights reduces the discrepancy and the objective-conditioned certificate as target probability approaches one; the geometry-only bound remains fixed.

\subsection{Class-weight rearrangement at fixed target probability}
\label{app:fixed-target-permutation}
For each of the 5,120 saved states, let $P$ be its exact reference probability vector, $y$ its target, and $j_1,\ldots,j_{C-1}$ its non-target classes in their stored order. Define $w_a=P_{j_a}/(1-P_y)$. Each prescribed permutation $\sigma$ produces
\[
P_y^{(\sigma)}=\tfrac12,\qquad
P_{j_a}^{(\sigma)}=\tfrac12 w_{\sigma(a)}.
\]
The target probability $0.5$ provides a common nonsaturated control condition. The classwise weighting mechanism in Eq.~\eqref{eq:gradient-mechanism-main} applies more generally across target probabilities.

The five permutations are identity, a cyclic shift by one position, a cyclic shift by $\lfloor(C-1)/2\rfloor$ positions, reversal, and a pseudorandom permutation generated with seed 3407. They are fixed before comparing downstream errors.

A query-specific additive class bias realizes each vector: if $q_j$ is the original exact logit, add $\beta_j^{(\sigma)}=\log P_j^{(\sigma)}-q_j$ to both exact and approximate logits. With $e_j=\widehat q_j-q_j$, their new probabilities are $P^{(\sigma)}$ and $\pi_j^{(\sigma)}\propto P_j^{(\sigma)}\exp(e_j)$. Treating the added biases as constants during feature differentiation preserves the exact and approximate class-gradient vectors. Thus the intervention changes only the correspondence between probability weights and the fixed classwise errors and directions. Dimension, temperature, feature norm, endpoints, class directions, score errors, ratio errors, and primitive envelopes remain unchanged.

The non-target probability multiset, reference entropy, and exact top-two logit margin
$\log[(1/2)/\max_{j\ne y}P_j^{(\sigma)}]$ are invariant across permutations. These invariants refer to the exact distribution; the approximate target probability is allowed to respond to the unchanged classwise numerical errors. This controlled modification isolates the weighting mechanism at each saved state.

Table~\ref{tab:permutation-spread} compares relative variation with absolute error ranges. Large fold changes can arise from very small minimum errors, whereas Table~\ref{tab:permutation-budgets} measures the consequence for actual loss and gradient budgets. Together, the tables distinguish a change in learning error from a change in whether that error meets a specified requirement.

\begin{table}[ht]
\centering
\caption{Class assignments change learning errors at fixed probability summaries. Fold change is the largest error divided by the smallest error for each state; absolute range is their difference.}
\label{tab:permutation-spread}
\begin{tabular}{lrr}
\toprule
Summary & Loss & Feature gradient\\
\midrule
Fold change: median & 3.18 & 3.28\\
Fold change: maximum & $1.06\!\times\!10^{7}$ & $1.16\!\times\!10^{7}$\\
Median absolute range & $2.36\!\times\!10^{-9}$ & $9.09\!\times\!10^{-9}$\\
Maximum absolute range & $1.78\!\times\!10^{-7}$ & $8.55\!\times\!10^{-7}$\\
\bottomrule
\end{tabular}
\end{table}

\begin{table}[ht]
\centering
\caption{Actual numerical adequacy under class-weight rearrangement. Counts give the states for which the five assignments yield different adequacy decisions. The joint requirement applies the same budget to absolute loss error and Euclidean feature-gradient error.}
\label{tab:permutation-budgets}
\begin{tabular}{lrrr}
\toprule
Budget & Loss & Gradient & Joint\\
\midrule
$10^{-4}$ & 0 & 0 & 0\\
$10^{-5}$ & 0 & 0 & 0\\
$10^{-6}$ & 0 & 0 & 0\\
$10^{-7}$ & 56 & 1014 & 1001\\
$10^{-8}$ & 1150 & 526 & 521\\
$10^{-9}$ & 571 & 148 & 138\\
\bottomrule
\end{tabular}
\end{table}

The five assignments per query preserve the reference probability summaries while changing their correspondence with numerical errors. Their signed loss changes lie within the rearrangement extrema of Proposition~\ref{prop:ce-coupling}. Reference evaluation uses the common geometry described in Appendix~\ref{app:numerics-reference}.

%% file: sections/appendix_primitive.tex
\section{The Paired vMF Instance}
\label{sec:primitive}
\label{sec:primitive-pair}
Put $s(x)=\sqrt{\nu^2+x^2}$. The realization used in the main experiments is
\begin{align}
A_{2,\nu}(x)&=\frac{x}{\nu+s(x)}-\frac{x}{2s(x)^2}
+\frac{x(4\nu^2-x^2)}{8s(x)^5},\\
F_{2,\nu}(x)&=s(x)-\nu\log(\nu+s(x))-\frac12\log s(x)
+\frac{1}{8s(x)}-\frac{5\nu^2}{24s(x)^3}.
\end{align}
Direct differentiation gives $F_{2,\nu}'=A_{2,\nu}$ and $A_{2,\nu}(0)=0$. Additive constants in the potential cancel from endpoint differences. The construction uses the modified-Bessel asymptotic structure \citep{olver1954asymptotic,temme1996special}; its role here is to provide a concrete paired realization with a provable allowance.
\begin{theorem}[Uniform primitive allowance]
\label{thm:a2}
For every $\nu\ge10/9$ and $x\ge0$,
\[
|R_\nu(x)-A_{2,\nu}(x)|\le b_\nu,\qquad b_\nu=\frac{1}{4\nu^3}.
\]
\end{theorem}
Appendix~\ref{app:proof} proves this bound. Thus $a_j=b_\nu$ and $d_j=b_\nu|r_j-k_j|$ are valid inputs to the main-text contract.

\paragraph{Refined allowance interface.}
\label{sec:primitive-state}
Let $m=2\nu+2$ and
\[
D_\nu(t)=\nu^{-3}\left[(1+(t/\nu)^2)^{-3/2}
+\frac{1}{4\nu}(1+(t/\nu)^2)^{-2}\right].
\]
For $0\le\alpha\le1$, a valid pointwise envelope is
\[
B_\alpha(x)=\frac{x}{m}\{D_\nu(0)\alpha^m+D_\nu(\alpha x)(1-\alpha^m)\}.
\]
Taking its minimum with $b_\nu$, or with other valid choices of $\alpha$, preserves validity. Appendix~\ref{app:realization-state} derives the envelope and Appendix~\ref{app:realization-score} gives valid interval bounds for score integration. These certificates all describe identical $F_2/A_2$ outputs. The stable endpoint formula is given in Appendix~\ref{app:realization-endpoint}.

\paragraph{Special-function context.}
Classical algorithms evaluate modified Bessel functions and their ratios through recurrence, asymptotics, and region-dependent numerical methods \citep{amos1974computation,gil2007numerical}. Analytic ratio inequalities provide another source of usable allowances \citep{hornik2013amos,ruizantolin2016bounds,segura2023simple}. Here these tools supply the primitive-level premise; the state-conditioned tolerance additionally depends on its propagation through the learning objective.

%% file: sections/appendix_foundation.tex
\section{Mathematical Foundation of the Analytic Primitive}
\label{app:proof}

This appendix proves the uniform ratio certificate for the analytic
vMF primitive in Section~\ref{sec:interface}. We first derive a preliminary
$\nu^{-3}$ enclosure directly from the Riccati equation, then sharpen the
barrier argument to obtain the factor-$1/4$ bound in
Theorem~\ref{thm:a2}. The proof supplies the uniform allowance used by the inexpensive certificate.

\subsection{Riccati identity and a preliminary ratio bound}
\label{app:proof-preliminary}

Set
\[
\varepsilon=\nu^{-1},
\qquad
x=\nu z,
\]
and define
\begin{equation}
H_\varepsilon(z)
=
\frac{I_{\nu+1}(\nu z)}{I_\nu(\nu z)}.
\label{eq:app-H}
\end{equation}
The adjacent-order modified-Bessel ratio satisfies the Riccati equation
\begin{equation}
\varepsilon H_\varepsilon'(z)
=
1
-
\frac{2+\varepsilon}{z}H_\varepsilon(z)
-
H_\varepsilon(z)^2.
\label{eq:riccati-app}
\end{equation}

Let
\[
S=\sqrt{1+z^2}
\]
and define
\begin{equation}
\begin{aligned}
r_0(z)
&=
\frac{z}{1+S},
\\
r_1(z)
&=
-\frac{z}{2S^2},
\\
r_2(z)
&=
\frac{z(4-z^2)}{8S^5},
\\
a_\varepsilon(z)
&=
r_0(z)
+
\varepsilon r_1(z)
+
\varepsilon^2 r_2(z).
\end{aligned}
\label{eq:riccati-candidate}
\end{equation}
Substituting $x=\nu z$ shows that
\[
a_\varepsilon(z)=A_{2,\nu}(x).
\]

We evaluate how accurately $a_\varepsilon$ satisfies the exact Riccati
equation. Define the residual
\begin{equation}
\mathcal E
=
1
-
\frac{2+\varepsilon}{z}a_\varepsilon
-
a_\varepsilon^2
-
\varepsilon a_\varepsilon'.
\label{eq:app-residual-def}
\end{equation}
Direct simplification gives the exact identity
\begin{equation}
\begin{aligned}
\mathcal E
={}&
-\varepsilon^3
\frac{z^4-10z^2+4}
     {4(1+z^2)^{7/2}}
\\
&-
\varepsilon^4
\frac{z^2(z^2-4)^2}
     {64(1+z^2)^5}.
\end{aligned}
\label{eq:exact-residual}
\end{equation}
For $u=z^2\ge0$,
\begin{equation}
|u^2-10u+4|
\le
4(1+u)^2,
\qquad
u(u-4)^2
\le
16(1+u)^3.
\label{eq:app-poly-bounds}
\end{equation}
Applying these inequalities to
Eq.~\eqref{eq:exact-residual} yields
\begin{equation}
|\mathcal E(z,\varepsilon)|
\le
\varepsilon^3
\left(1+\frac{\varepsilon}{4}\right).
\label{eq:residual-bound}
\end{equation}

Let
\[
e(z)
=
H_\varepsilon(z)-a_\varepsilon(z).
\]
Subtracting the candidate equation from
Eq.~\eqref{eq:riccati-app} gives
\begin{equation}
\varepsilon e'
=
\mathcal E
-
Ke
-
e^2,
\label{eq:error-ode}
\end{equation}
where
\begin{equation}
\begin{aligned}
K
&=
\frac{2+\varepsilon}{z}
+
2a_\varepsilon
\\
&=
\frac{2S}{z}
+
\frac{\varepsilon}{zS^2}
+
2\varepsilon^2r_2(z).
\end{aligned}
\label{eq:app-K}
\end{equation}

When $0<z\le2$, we have $r_2(z)\ge0$. For $z\ge2$,
\begin{equation}
-r_2(z)
=
\frac{z(z^2-4)}{8S^5}
\le
\frac{1}{40}.
\label{eq:app-r2-bound}
\end{equation}
Therefore, whenever
$0<\varepsilon\le9/10$,
\begin{equation}
K
\ge
2-\frac{\varepsilon^2}{20}
\ge
\frac{3919}{2000}.
\label{eq:k-bound}
\end{equation}

We first choose the constant barrier
\[
b=\varepsilon^3.
\]
At the upper boundary $e=b$,
Eqs.~\eqref{eq:residual-bound} and
\eqref{eq:k-bound} imply
\[
\mathcal E-Kb-b^2<0.
\]
Thus the vector field in Eq.~\eqref{eq:error-ode}
points inward at $e=b$.

At the lower boundary $e=-b$, the right-hand side of
Eq.~\eqref{eq:error-ode} is bounded below by
\begin{equation}
\varepsilon^3
\left(
-\frac{49}{40}
+
\frac{3919}{2000}
-
\frac{729}{1000}
\right)
=
\frac{11}{2000}\varepsilon^3
>
0.
\label{eq:lower-margin}
\end{equation}
Hence the vector field also points inward at $e=-b$.

It remains to initialize the barrier argument at the singular endpoint.
The small-$z$ expansions are
\begin{equation}
H_\varepsilon(z)
=
\frac{z}{2(1+\varepsilon)}
+
O(z^3)
\label{eq:app-small-H}
\end{equation}
and
\begin{equation}
a_\varepsilon(z)
=
\frac{z}{2}
\left(
1-\varepsilon+\varepsilon^2
\right)
+
O(z^3).
\label{eq:app-small-a}
\end{equation}
Subtracting gives
\begin{equation}
e(z)
=
-\frac{\varepsilon^3}
       {2(1+\varepsilon)}
z
+
O(z^3).
\label{eq:app-small-e}
\end{equation}
Thus $e$ begins strictly inside $[-b,b]$ for sufficiently small positive
$z$.

If a first contact with the upper barrier existed, it would require
$e'\ge0$ at that point, contradicting the strict inward direction proved
above. A first contact with the lower barrier would similarly require
$e'\le0$ and contradict the opposite inward direction. Consequently,
\begin{equation}
|e(z)|<\varepsilon^3,
\qquad
z>0.
\label{eq:app-preliminary}
\end{equation}
Both $H_\varepsilon$ and $a_\varepsilon$ extend continuously to $z=0$
with value zero, so the corresponding non-strict bound holds at the
origin. Since
\[
\varepsilon\le\frac{9}{10}
\qquad\Longleftrightarrow\qquad
\nu\ge\frac{10}{9},
\]
we obtain the preliminary all-axis enclosure
\[
|R_\nu(x)-A_{2,\nu}(x)|
\le
\nu^{-3}.
\]
The next subsection sharpens this constant to the factor $1/4$ stated in
Theorem~\ref{thm:a2}.

\subsection{Refined uniform certificate}
\label{app:refined}

The preliminary proof replaces the concentration-dependent residual by a
single uniform upper bound. Retaining its denominators yields a sharper
comparison. With the notation above,
\begin{equation}
|\mathcal E|
\le
\varepsilon^3
\left[
S^{-3}
+
\frac{\varepsilon}{4}S^{-4}
\right].
\label{eq:app-refined-residual}
\end{equation}
Moreover,
\begin{equation}
K
\ge
\kappa\frac{2S}{z},
\qquad
\kappa=\frac{3919}{4000}.
\label{eq:app-refined-K}
\end{equation}
Indeed, Eq.~\eqref{eq:app-K} and
Eq.~\eqref{eq:app-r2-bound} give
\[
K
\ge
\frac{2S}{z}
-
\frac{\varepsilon^2}{20},
\]
while
\[
\frac{2S}{z}\ge2
\qquad
\text{and}
\qquad
\varepsilon\le\frac{9}{10}.
\]

The relevant rational functions satisfy
\begin{equation}
\sup_{z\ge0}
\frac{z}{2S^4}
=
\frac{9}{32\sqrt3},
\qquad
\sup_{z\ge0}
\frac{z}{2S^5}
=
\frac{8}{25\sqrt5}.
\label{eq:app-rational-suprema}
\end{equation}
Combining
Eqs.~\eqref{eq:app-refined-residual}--\eqref{eq:app-rational-suprema}
gives
\begin{equation}
\frac{|\mathcal E|}
     {K\varepsilon^3}
\le
\kappa^{-1}
\left[
\frac{9}{32\sqrt3}
+
\frac{9}{40}
\frac{8}{25\sqrt5}
\right]
<
\frac15.
\label{eq:app-residual-over-K}
\end{equation}
The final strict inequality already follows from the rational lower bounds
\[
\sqrt3>\frac{1732}{1000},
\qquad
\sqrt5>\frac{2236}{1000}.
\]

We now choose the refined constant barrier
\begin{equation}
B=\frac{\varepsilon^3}{4}.
\label{eq:app-quarter-barrier}
\end{equation}
At $e=B$,
Eq.~\eqref{eq:app-residual-over-K} gives
\[
\mathcal E-KB-B^2<0.
\]
At $e=-B$, the right-hand side of
Eq.~\eqref{eq:error-ode} is bounded below by
\begin{equation}
K\varepsilon^3
\left(
\frac14
-
\frac15
-
\frac{\varepsilon^3}{16K}
\right).
\label{eq:app-quarter-lower}
\end{equation}
Furthermore,
\begin{equation}
\frac{\varepsilon^3}{16K}
\le
\frac{1458}{62704}
<
\frac1{20},
\label{eq:app-quarter-last}
\end{equation}
so Eq.~\eqref{eq:app-quarter-lower} is strictly positive.

The same small-$z$ initialization from
Eq.~\eqref{eq:app-small-e} and the same first-contact argument used in
Appendix~\ref{app:proof-preliminary} therefore yield
\begin{equation}
|H_\varepsilon(z)-a_\varepsilon(z)|
\le
\frac{\varepsilon^3}{4},
\qquad
z\ge0.
\label{eq:app-quarter-result}
\end{equation}
Returning to
$x=\nu z$ and $\varepsilon=\nu^{-1}$ gives
\begin{equation}
|R_\nu(x)-A_{2,\nu}(x)|
\le
\frac{1}{4\nu^3},
\qquad
\nu\ge\frac{10}{9},
\quad
x\ge0,
\end{equation}
which proves Theorem~\ref{thm:a2}.

%% file: sections/appendix_learning.tex
\section{Proofs of the Learning-Level Contracts}
\label{app:learning}
\subsection{Scores and cross-entropy}
\label{app:learning-score}
Pairing gives
\[
(q_j-\widehat q_j)=\int_{k_j}^{r_j}(R_\nu(x)-A(x))\,dx,
\]
so the integrated primitive envelope bounds its absolute value by $d_j$. Since log-sum-exp is coordinatewise increasing and invariant to a common shift up to that shift,
\[
\LSE(\widehat q)-d\le\LSE(q)\le\LSE(\widehat q)+d.
\]
Adding the target-score error proves $|L_y-\widehat L_y|\le d+d_y$.

\subsection{Probability TV and prediction stability}
\label{app:learning-probability}
Put $e_j=\widehat q_j-q_j$, so $|e_j|\le d$. Under the exact distribution $P$, let the positive random variable $X$ take values $\exp(e_j)$. Then $X\in[a,b]=[e^{-d},e^d]$, and
\[
\TV(P,\pi)=\frac{\mathbb E_P|X-\mathbb E_PX|}{2\mathbb E_PX}.
\]
For $m=\mathbb E_PX$, convexity bounds the numerator by the chord joining the endpoints:
\[
\mathbb E_P|X-m|\le\frac{2(b-m)(m-a)}{b-a}.
\]
Maximizing $(b-m)(m-a)/(m(b-a))$ over $m\in[a,b]$ gives
\[
\TV(P,\pi)\le\frac{\sqrt b-\sqrt a}{\sqrt b+\sqrt a}=\tanh(d/2).
\]
For $d=0$ the distributions coincide. If $j_\star$ is the approximate winner, then
$q_{j_\star}-q_j\ge\widehat q_{j_\star}-\widehat q_j-2d$.
An approximate margin greater than $2d$ therefore certifies the winner. Interchanging exact and approximate scores proves the exact-margin version.

\subsection{Feature-gradient propagation}
\label{app:learning-gradient}
All class statistics, biases, and $\tau$ are held fixed. Define
\[
h_j=R_\nu(r_j)v_j/r_j,\quad \widehat h_j=A(r_j)v_j/r_j,
\quad \nabla_f q_j=h_j/\tau,\quad\nabla_f\widehat q_j=\widehat h_j/\tau,
\]
with continuous zero extensions when $r_j=0$. Because $0\le R_\nu\le1$, $\norm{h_j}\le1$ and $\norm{\widehat h_j-h_j}\le a_j$. Subtracting the two cross-entropy gradients gives Eq.~\eqref{eq:gradient-mechanism-main}. The probability-change term has norm at most $2\TV(P,\pi)$; the primitive-error term has norm at most
\[
\sum_{j\ne y}\pi_j a_j+(1-\pi_y)a_y\le\max_j a_j+a_y.
\]
Together with the TV bound, this proves Proposition~\ref{prop:contrastive-contract}.

\subsection{Objective-conditioned intervals}
\label{app:probability-conditioned}
The exact probability can be written in terms of approximate quantities as
\[
P_j=\frac{\pi_j}{D_j(e)},\qquad
D_j(e)=\pi_j+\sum_{i\ne j}\pi_i\exp(e_j-e_i).
\]
The self-term has no perturbation. Since $|e_j-e_i|\le d_j+d_i$,
\[
0<1+S_j^-\le D_j(e)\le1+S_j^+.
\]
This proves the main-text probability intervals, with the upper endpoint optionally clipped at one. Also $L_y-\widehat L_y=\log D_y(e)$, so its absolute value is at most $B_L$. The two loss endpoints are attained over the independent error box by choosing opposite target and non-target signs. They need not be jointly realizable by a single paired primitive.

The coordinate intervals give
\[
\TV(P,\pi)\le\frac12\sum_j\max(\pi_j-\ell_j,u_j-\pi_j).
\]
Combining this with the uniform TV bound proves $B_P$. Substituting $B_P$ and the probability-weighted primitive-error term into the gradient decomposition proves $B_g$.

These bounds dominate the base contract: the loss denominator has logarithm in $[-d_y-d,d_y+d]$, the TV bound explicitly takes a minimum with the base bound, and the weighted ratio term is at most $\max_j a_j+a_y$. This proves Proposition~\ref{prop:objective-conditioned-contract}. The coordinate intervals need not be jointly attainable; the resulting TV and gradient bounds are sufficient rather than sharp in general.

\paragraph{Computation.}
Let $m_i^\pm=\operatorname{expm1}(\pm d_i)$. Then
\[
S_j^\pm=m_j^\pm(1-\pi_j)+(1+m_j^\pm)
\left(\sum_i\pi_i m_i^\pm-\pi_jm_j^\pm\right).
\]
This reduces the mathematical construction to linear rather than quadratic work in class count. Log-domain exclusive sums are used for wide allowance ranges during inversion.

\paragraph{Fixed-state scope.}
A common target-class bias changes probabilities while leaving every primitive error and allowance fixed. As that bias tends to infinity, the exact and approximate target probabilities tend to one; the loss and feature-gradient errors and their objective-conditioned bounds tend to zero. Changing $f=\tau u$ at fixed $u$ instead preserves probabilities and all primitive inputs, but changes the named feature-gradient scale by $1/\tau$. These are the distinct interventions in Section~\ref{sec:exp-conditioning}.

\subsection{Proof of class-weight--error coupling}
\label{app:ce-coupling-proof}
Subtracting $L_y(q)=\log\sum_j\exp(q_j)-q_y$ from $L_y(q+e)$ gives Eq.~\eqref{eq:ce-error-identity}. Subtracting $e_y$ inside the logarithm rewrites it as $\log[t+\sum_{j\ne y}P_j\exp(e_j-e_y)]$. For $p_i\le p_k$ and $z_i\le z_k$, the exchange identity
\[
(p_iz_i+p_kz_k)-(p_iz_k+p_kz_i)=(p_k-p_i)(z_k-z_i)\ge0
\]
shows that matched ordering maximizes the weighted sum and reverse ordering minimizes it. Apply this classical rearrangement argument to $z_i=\exp(a_i-e_y)$. Adding $t$ and taking the increasing logarithm preserves the extrema, proving Proposition~\ref{prop:ce-coupling}. Every signed change lies between the two attained endpoints; its largest absolute value is therefore the larger endpoint magnitude. The finite set of pairings need not attain zero even when $D_-<0<D_+$. Common shifts of all errors leave the identity unchanged. For $C\ge3$, if neither non-target multiset is constant, the two extrema differ. These extrema characterize signed loss changes. Ordering gradient-error norms would additionally require accounting for the class directions in the gradient decomposition.

%% file: sections/appendix_objectives.tex
\section{A Second vMF Objective}
\label{app:objectives}
The propagation interface also applies to supervised probabilistic vMF embeddings \citep{Scott_2021_ICCV}, where the primitive enters both a log-normalizer difference and a ratio-dependent attraction term. Write $z=a\mu_z$, $w_j=k_j\mu_j$, and fix common unit samples $s$ and inverse temperature $\beta>0$. The objective is
\[
L=\mathbb E_s\LSE_j\{\Phi_\nu(\norm{w_j+\beta s})-\Phi_\nu(k_j)\}
-\beta R_\nu(a)R_\nu(k_y)\mu_z^\top\mu_y.
\]
Replace $\Phi_\nu,R_\nu$ by a paired $F,A$ with uniform ratio allowance $b$. The reverse triangle inequality gives $|\norm{w_j+\beta s}-k_j|\le\beta$, so the normalizer-difference score error is at most $\beta b$. Log-sum-exp therefore changes by at most $\beta b$ for each fixed sample. Since $0\le R_\nu\le1$ and $|A-R_\nu|\le b$,
\[
|A(a)A(k_y)-R_\nu(a)R_\nu(k_y)|\le2b+b^2.
\]
Consequently,
\[
|\widehat L-L|\le\beta(3b+b^2).
\]
The bound holds for a common finite sample average and after expectation over a common distribution. Class-specific endpoint and ratio allowances can replace these uniform terms. The same primitive-allowance interface thus supports the distinct propagation algebra of the supervised vMF objective.

%% file: sections/appendix_realization.tex
\section{Numerical Realization and Certificate Construction}
\label{app:realization}

This appendix derives the state-dependent primitive envelope, constructs score allowances, and gives the stable endpoint representation used by the paired surrogate.

\subsection{State-dependent residual envelope}
\label{app:realization-state}

Let
\[
e(x)
=
R_\nu(x)-A_{2,\nu}(x)
\]
denote the exact pointwise ratio error. Returning from the rescaled
Riccati analysis of Appendix~\ref{app:proof} to the original coordinate,
$e$ satisfies
\begin{equation}
e'
+
\left[
\frac{2\nu+1}{x}
+
R_\nu(x)
+
A_{2,\nu}(x)
\right]
e
=
\mathcal E(x),
\label{eq:app-state-error-ode}
\end{equation}
where the residual admits the concentration-dependent bound summarized
below.

Define
\begin{equation}
m=2\nu+2
\label{eq:app-state-m}
\end{equation}
and
\begin{equation}
D_\nu(t)
=
\nu^{-3}
\left[
\left(
1+(t/\nu)^2
\right)^{-3/2}
+
\frac{1}{4\nu}
\left(
1+(t/\nu)^2
\right)^{-2}
\right].
\label{eq:app-Dnu}
\end{equation}

The coefficient multiplying $e$ in
Eq.~\eqref{eq:app-state-error-ode} is bounded below by
\[
\frac{2\nu+1}{x}.
\]
For the exact ratio, nonnegativity is immediate:
\[
R_\nu(x)\ge0.
\]
For $A_{2,\nu}$, use the rescaled coordinate
$z=x/\nu$ and the notation of
Appendix~\ref{app:proof}. When $z\le2$,
\[
r_2(z)\ge0
\]
and
\begin{equation}
r_0+\varepsilon r_1
\ge
r_0
\left(
1-\frac{\varepsilon}{S}
\right).
\label{eq:app-A2-nonnegative-small}
\end{equation}
When $z\ge2$,
\begin{equation}
\frac{|r_2|}{r_0}
\le
\frac{1}{4S^2},
\qquad
S\ge\sqrt5.
\label{eq:app-A2-nonnegative-large}
\end{equation}
These inequalities imply
\[
A_{2,\nu}(x)\ge0
\]
throughout the theorem domain. Hence the coefficient of $e$ is indeed at
least $(2\nu+1)/x$.

Using the continuous endpoint condition
\[
e(0)=0
\]
and variation of constants in
Eq.~\eqref{eq:app-state-error-ode} gives
\begin{equation}
|e(x)|
\le
x^{-(m-1)}
\int_0^x
t^{m-1}
D_\nu(t)\,dt.
\label{eq:app-state-integral-envelope}
\end{equation}

The function $D_\nu$ is decreasing. For any
\[
\alpha\in[0,1],
\]
split the integral at $\alpha x$. On the first interval,
\[
D_\nu(t)\le D_\nu(0),
\]
while on the second,
\[
D_\nu(t)\le D_\nu(\alpha x).
\]
Therefore
\begin{align}
\int_0^x
t^{m-1}D_\nu(t)\,dt
\le{}&
D_\nu(0)
\int_0^{\alpha x}
t^{m-1}\,dt
\nonumber\\
&+
D_\nu(\alpha x)
\int_{\alpha x}^{x}
t^{m-1}\,dt.
\label{eq:app-state-split}
\end{align}
Evaluating the two monomial integrals and multiplying by
$x^{-(m-1)}$ yields the pointwise bound
\begin{equation}
B_\alpha(x)
=
\frac{x}{m}
\left\{
D_\nu(0)\alpha^m
+
D_\nu(\alpha x)
\left(
1-\alpha^m
\right)
\right\}.
\label{eq:app-Balpha}
\end{equation}
Thus, for every admissible $\alpha$,
\begin{equation}
|R_\nu(x)-A_{2,\nu}(x)|
\le
B_\alpha(x).
\label{eq:app-Balpha-valid}
\end{equation}

Any minimum over valid choices remains a valid certificate. In particular,
the implementation used in the experiments takes the minimum of the uniform
quarter allowance and six state-dependent candidates with
\begin{equation}
\alpha
=
\exp(-c/m),
\qquad
c\in\{1,2,4,8,16,32\}.
\label{eq:app-alpha-grid}
\end{equation}
Each candidate is valid separately; their minimum tightens the certificate.

\subsection{Score certificates from pointwise envelopes}
\label{app:realization-score}

For an endpoint interval
\[
[l,h],
\qquad
0\le l\le h,
\]
a valid pointwise envelope $B(x)$ on the ratio error yields
\begin{equation}
\left|
\int_l^h
\left[
R_\nu(x)-A_{2,\nu}(x)
\right]dx
\right|
\le
\int_l^h
B(x)\,dx.
\label{eq:app-score-envelope-integral}
\end{equation}

For the state-dependent candidate
$B_\alpha(x)$ in
Eq.~\eqref{eq:app-Balpha}, a simple upper-sum construction is obtained by
partitioning the endpoint interval into panels. On a panel
$[l_p,h_p]$, we use
\[
\frac{x}{m}
\le
\frac{h_p}{m}
\]
and, because $D_\nu$ is decreasing,
\[
D_\nu(\alpha x)
\le
D_\nu(\alpha l_p).
\]
Hence
\begin{equation}
B_\alpha(x)
\le
\frac{h_p}{m}
\left\{
D_\nu(0)\alpha^m
+
D_\nu(\alpha l_p)
\left(
1-\alpha^m
\right)
\right\}
\label{eq:app-panel-envelope}
\end{equation}
for every $x\in[l_p,h_p]$.

Multiplying the right-hand side by the panel width and summing over all panels
therefore produces an exact-arithmetic upper sum for the score error. The
experiments use 16 equal panels; any partition with valid pointwise envelopes preserves the guarantee.

\subsection{Stable endpoint realization}
\label{app:realization-endpoint}

The vMF objectives use potential differences
\[
F_{2,\nu}(r)-F_{2,\nu}(k)
\]
rather than isolated potential values. Direct subtraction of two nearly equal
floating-point evaluations can lose substantial relative accuracy.

Let
\[
d=r^2-k^2,
\qquad
s_0=s(k),
\qquad
s_1=s(r),
\]
and define the rationalized radius difference
\begin{equation}
\Delta_s
=
\frac{d}{s_1+s_0}.
\label{eq:app-delta-s}
\end{equation}
Instead of subtracting two separately evaluated potential values, compute
\begin{align}
\widehat q(r,k)
={}&
\Delta_s
-
\nu
\operatorname{log1p}
\left(
\frac{\Delta_s}{\nu+s_0}
\right)
-
\frac12
\operatorname{log1p}
\left(
\frac{\Delta_s}{s_0}
\right)
\nonumber\\
&-
\frac{\Delta_s}{8s_1s_0}
+
\frac{
5\nu^2\Delta_s
\left(
s_1^2+s_1s_0+s_0^2
\right)
}{
24s_1^3s_0^3
}.
\label{eq:stable-f2-app}
\end{align}
In exact arithmetic,
\begin{equation}
\widehat q(r,k)
=
F_{2,\nu}(r)-F_{2,\nu}(k).
\label{eq:app-stable-equivalence}
\end{equation}

Equation~\eqref{eq:stable-f2-app} is an algebraically equivalent evaluation of
the same paired surrogate and preserves its exact-arithmetic certificates.

When the application supplies the squared-radius difference $d$ directly,
Eq.~\eqref{eq:app-delta-s} avoids first constructing and subtracting two
nearly identical rounded radii. With rounded endpoint radii as inputs, the attainable accuracy is also
limited by information lost during their construction.

%% file: sections/appendix_numerics.tex
\section{Reference Evaluation and Numerical Scope}
\label{app:numerics}
\subsection{Saved-state provenance}
\label{app:numerics-states}
The saved queries come from ProCo, PATT, and a vMF contrastive setting with its auxiliary classification term disabled \citep{du2024proco,he2025patt}, with CIFAR, ImageNet-LT, and iNaturalist class banks \citep{krizhevsky2009learning,deng2009imagenet,vanhorn2018inaturalist}. Each query retains its label, complete class statistics, prior, feature geometry, and temperature. Queries sharing a source batch or training run are not independent replications.

\subsection{High-precision reference at shared geometry}
\label{app:numerics-reference}
\label{app:numerics-objective-replay}
Each radius is reconstructed once in FP64 and supplied unchanged to the Bessel reference, paired realization, and primitive envelopes. Scalar scores, ratios, and probabilities are evaluated with 80-digit \texttt{mpmath} arithmetic \citep{mpmath2026}. Gradient coefficient differences are formed at high precision before contraction with the common FP64 direction vectors. This shared geometry isolates primitive approximation from differences in endpoint reconstruction.

Certificate inversion uses FP64 log-domain calculations, with the propagated bounds and inverse boundaries evaluated at high precision for the reference comparison. Stable formulas preserve small perturbations without subtracting nearly equal probabilities \citep{blanchard2021softmax}. No bound violation is observed in the saved-query comparison, and all predictions satisfying the margin certificate are preserved. The analytic proof establishes the mathematical guarantee; these comparisons assess its numerical realization.

\subsection{Supplementary actual-response check under a fixed perturbation family}
\label{app:actual-response}
\label{app:empirical-sensitivity}
The perturbation family $F_b=F_2-bH_\nu$ is fixed before observing the responses, with $H_\nu(x)=\sqrt{\nu^2+x^2}-\nu$. Its derivative perturbation is bounded by $b$ and vanishes at zero. We keep the saved geometry, biases, and targets fixed and recompute perturbed probabilities and feature gradients. Stable probability differences and shared direction vectors isolate the induced gradient discrepancy. The unchanged paired potential $F_2$ serves as the reference for this response comparison.

On the sampled endpoints, $A_2$ and $H_\nu'$ lie in $[0,1]$. For $0\le b\le1$, the perturbed radial derivative therefore has magnitude at most one. Taking the unchanged $F_2$ realization as the approximate side of the gradient decomposition makes the same fixed-probability contract applicable, with $d_j=b|r_j-k_j|$ and $a_j=b$. This preserves the saved certified threshold while measuring the actual gradient difference independently.

We locate the first budget crossing resolved by a scan on $0\le b\le1$ and refine it locally, then compare it with the same query's gradient certificate. The resulting threshold is defined at the scan's resolution; a nonmonotone response may have narrower crossings between scan points.

Actual gradient-response thresholds span multiple orders of magnitude: 5.38 at budget $10^{-4}$ and 6.35 at $10^{-5}$, with a span of 5.70 at $10^{-5}$ even within the fixed $p=512$, $\tau=0.1$ group. The sufficient certificate is typically more conservative for this fixed perturbation family; the median detected response threshold is 24.61 times the gradient certificate at $10^{-5}$.

\subsection{Stable realization and the floating-point boundary}
\label{app:numerics-endpoint}
Automatic differentiation through the paired potential agrees with its explicit derivative at the precision expected from the tested arithmetic \citep{griewank2008evaluating,baydin2018autodiff}. Near coincident endpoints, the stable expression in Appendix~\ref{app:realization-endpoint} avoids the cancellation seen in direct subtraction. Those comparisons use identical stored squared-gap inputs and differentiation coordinates, separating realization error from Bessel-approximation error.

The mathematical allowance controls the exact-arithmetic surrogate. Floating-point endpoint construction, transcendental evaluation, reductions, and rounding introduce separate errors \citep{goldberg1991floating,higham2002accuracy}. Our numerical checks support the tested implementation range; a complete execution certificate would additionally require justified rounding allowances or outward-rounded computation. In particular, a probability rounded to zero is not automatically an exact zero in the contract.